\documentclass[sigconf]{acmart}
\AtBeginDocument{%
  }

\copyrightyear{2025}
\acmYear{2025}
\setcopyright{acmlicensed}
\acmConference[CIKM '25]{Proceedings of the 34th ACM International Conference on Information and Knowledge Management}{November 10--14, 2025}{Seoul, Republic of Korea}
\acmBooktitle{Proceedings of the 34th ACM International Conference on Information and Knowledge Management (CIKM '25), November 10--14, 2025, Seoul, Republic of Korea}
 \acmDOI{10.1145/3746252.3761537}
\acmISBN{979-8-4007-2040-6/2025/11}
\usepackage{multirow}

\begin{document}

\title{Bridging the Gap Between Sparsity and Redundancy: A Dual-Decoding Framework with Global Context for Map Inference}

\author{Yudong Shen}
\email{51265903023@stu.ecnu.edu.cn}
\affiliation{%
  \institution{East China Normal University}
  \city{Shanghai}
  \country{China}
}

\author{Wenyu Wu}
\email{52275903019@stu.ecnu.edu.cn}
\affiliation{%
  \institution{East China Normal University}
  \city{Shanghai}
  \country{China}
}

\author{Jiali Mao}
\email{jlmao@dase.ecnu.edu.cn}
\affiliation{%
  \institution{East China Normal University}
  \city{Shanghai}
  \country{China}
}\authornote{corresponding author}

\author{Yixiao Tong}
\email{52265903007@stu.ecnu.edu.cn}
\affiliation{%
  \institution{East China Normal University}
  \city{Shanghai}
  \country{China}
}

\author{Guoping Liu}
\email{liuguoping@didiglobal.com}
\affiliation{%
  \institution{DiDi Chuxing}
  \city{Beijing}
  \country{China}
}

\author{Chaoya Wang}
\email{51265903103@stu.ecnu.edu.cn}
\affiliation{%
  \institution{East China Normal University}
  \city{Shanghai}
  \country{China}
}

\renewcommand{\shortauthors}{Yudong Shen, et al.}
\title[DGMap]{Bridging the Gap Between Sparsity and Redundancy: A Dual-Decoding Framework with Global Context for Map Inference}

\begin{abstract}
Trajectory data has become a key resource for automated map inference due to its low cost, broad coverage, and continuous availability. However, uneven trajectory density often leads to fragmented roads in sparse areas and redundant segments in dense regions, posing significant challenges for existing methods. To address these issues, we propose \emph{DGMap}, a dual-decoding framework with global context awareness, featuring \emph{Multi-scale Grid Encoding}, \emph{Mask-enhanced Keypoint Extraction}, and \emph{Global Context-aware Relation Prediction}. By integrating global semantic context with local geometric features, \emph{DGMap} improves keypoint detection accuracy to reduce road fragmentation in sparse-trajectory areas. 
Additionally, the \emph{Global Context-aware Relation Prediction} module suppresses false connections in dense-trajectory regions by modeling long-range trajectory patterns. Experimental results on three real-world datasets show that \emph{DGMap} outperforms state-of-the-art methods by 5\% in \emph{APLS}, with notable performance gains on trajectory data from the \emph{Didi Chuxing} platform.
\end{abstract}



\maketitle

\begin{figure}[t]
  \centering
  \includegraphics[width=\linewidth]{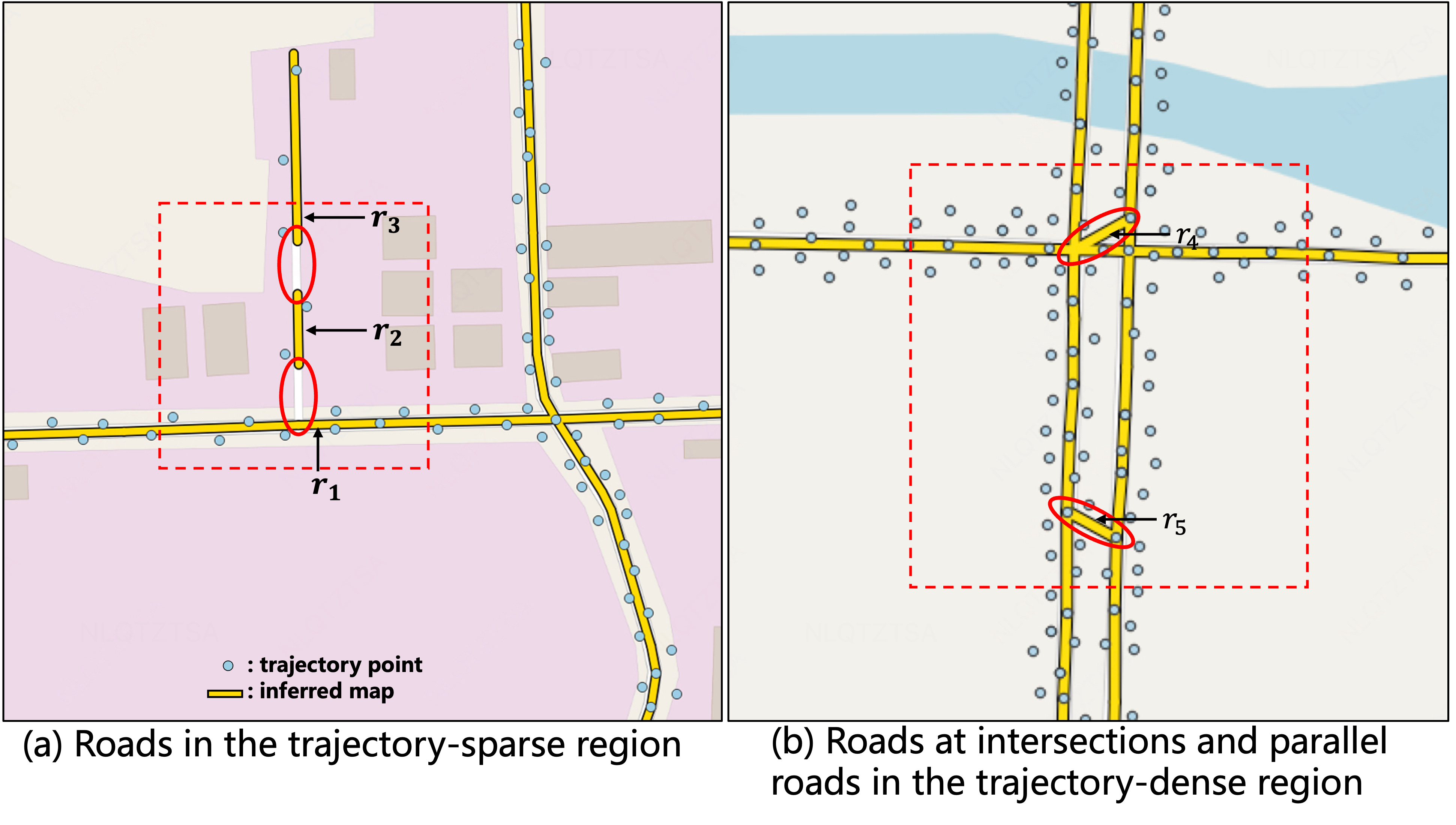}
  \caption{Illustration of Topological Discontinuities and Redundant Road Topologies. The yellow line segments represent the inferred road network, and the blue dots indicate the trajectory points.}
  \Description{Illustration of Topological Discontinuities and Redundant Road Topologies. The yellow line segments represent the inferred road network, and the blue dots indicate the trajectory points.}
\end{figure}

\section{Introduction}
Accurate digital maps serve as fundamental infrastructure for modern navigation and autonomous mobility systems, providing essential support for advancements in traffic safety, efficiency, and operational reliability. This critical role has positioned map inference as a prominent research domain in both academic and industrial settings. Vehicle trajectory data has emerged as a particularly valuable resource for map inference owing to its distinctive advantages: real-time data availability, comprehensive spatial coverage, and cost-efficient acquisition characteristics\cite{edelkamp2003route, cao2009gps, ahmed2012constructing, he2018roadrunner, stanojevic2018road, chao2020survey, shen2023sami, ruan2020learning, zhao2020automatic, eftelioglu2022ring, cao2015grarep}.

Conventional heuristic methods for trajectory-based map inference rely on manual parameter tuning, often leading to suboptimal accuracy due to improper settings\cite{ stanojevic2018robust, dey2017improved, karagiorgou2012vehicle}. In contrast, deep learning approaches have shown superior performance in complex urban environments by reducing heuristic-induced errors\cite{ruan2020learning}. These methods typically use an end-to-end framework that first identifies significant road features represented by \emph{keypoints} (e.g., structural nodes and geometric descriptors), and then establishes topological connections between \emph{keypoints} to reconstruct the complete road network\cite{he2022td}.

While deep learning methods achieve high accuracy in areas with sufficient trajectory density, their performance is hindered by the inherently uneven distribution of real-world trajectory data. This uneven distribution stems from variations in road network structures and regional functional zoning. For instance, urban arterial roads with heavy traffic accumulate dense trajectory data, whereas peripheral or sparsely populated areas suffer from insufficient coverage. Consequently, existing map inference techniques struggle to maintain spatial consistency, often producing incomplete or redundant road topologies.

\textbf{\textit{Challenge 1: Topological discontinuity commonly arise in inferred road regions with sparse trajectory coverage}}. In low-density trajectory regions, such as rural areas, the spatial sparsity of trajectory points fundamentally undermines the reliability of road network inference. Specifically, the average distance between trajectory points often exceeds the effective receptive field of deep neural networks, hindering the robust extraction of continuous geometric road features\cite{ruan2020learning}. This limitation manifests in the final reconstruction as topological discontinuities—particularly geometric fractures where roads are represented as piecewise linear segments instead of smooth, physically plausible curves—thereby degrading both the accuracy and completeness of the inferred map. As illustrated in Fig.1(a), the road segment $r_1$ is erroneously fragmented into disjoint segments $r_2$ and $r_3$ due to insufficient trajectory coverage, directly reflecting the model's reduced capacity to preserve spatial and topological coherence under sparse data conditions.

\textbf{\textit{Challenge 2: Redundant road topologies often emerge in areas where trajectory data are heavily intermingled.}} Road redundancy frequently occurs in parallel road and complex intersection scenarios due to trajectory data noise and spatial entanglement. Current solutions typically employ either heuristic distance thresholding or \emph{MLP-based} relation prediction to connect key points, but these methods focus solely on local point representations and fail to distinguish between valid connections and false shortcuts in trajectory-dense regions\cite{he2022td, yang2023topdig, hetang2024segment, yin2024towards, zorzi2022polyworld}. As shown in Fig. 1(b), this results in both spurious connections (e.g., $r_4$) from local pattern overfitting at intersections and erroneous parallel segments (e.g., $r_5$). The fundamental limitation stems from inadequate modeling of global topological relationships among trajectory points, leading to local overfitting in high-density areas and ultimately impairing the inferred road network's topological accuracy.

To address topology discontinuity from trajectory sparsity, we note that road region segmentation is more robust to sparse trajectories than keypoint detection, as it provides comprehensive road priors through grid-level dense prediction. Building on this insight, we develop a \emph{Mask-enhanced Keypoint Extraction} module (\emph{MeKE}) that synergistically combines road region segmentation with keypoint extraction, using attention-based feature interaction to enrich keypoint representations with regional features. Furthermore, to mitigate redundant links from dense trajectories, we propose a \emph{Global Context-aware Relation Prediction} module (\emph{GCRP}) that incorporates both keypoint and link-level global context via masked attention, significantly improving the discrimination between valid connections and redundant shortcuts.

In summary, we present \emph{DGMap}, a \emph{Dual-decoding framework with Global context} for map inference, comprising three key components: (i) a \emph{Multi-scale Grid Encoding} module that rasterizes road maps into grid cells and extracts hierarchical features using \emph{Deep Layer Aggregation} (\emph{DLA}); (ii) a \emph{Mask-enhanced Keypoint Extraction} module employing a dual-decoder architecture to boost keypoint detection accuracy through integrated road region representations; and (iii) a \emph{Global Context-aware Relation Prediction} module that leverages both keypoint-level and link-level global context for precise topological connection inference. Our main contributions are fourfold:
\begin{itemize}
    \item  We present \emph{DGMap}, a \emph{Dual-decoding framework with Global context} for map inference, which improves road topology accuracy through spatial consistency preservation by integrating three key components: \emph{Multi-scale Grid Encoding}, \emph{Mask-enhanced Keypoint Extraction}, and \emph{Global Context-aware Relation Prediction}.
    \item  To resolve topological discontinuities in sparse trajectories, we develop a \emph{Mask-enhanced Keypoint Extraction} module that leverages road region representations within a dual-decoding architecture to improve detection accuracy.
    \item To address the issue of redundant road topology generated by intermingled trajectory points, we put forward a \textit{Global Context-aware Relation Prediction} module by jointly modeling keypoint-level and link-level global contexts for precise connection inference.
    \item Experimental results on three real-world datasets demonstrate that our approach achieves consistent and significant improvements, outperforming state-of-the-art methods by an absolute 5\% in \emph{ALPS} on the \emph{DiDi} dataset.
\end{itemize}

\section{Related Work}
\subsection{Non-learning-based Map Inference Methods}
Unsupervised map inference methods primarily comprise three methodological paradigms: (i) \textit{Clustering-based} approaches construct road networks by spatially clustering trajectory points to identify intersections and connecting them using high-frequency trajectories. While effective in well-covered areas, these methods exhibit significant sensitivity to trajectory density variations, often generating redundant topologies in dense areas and fragmented connections in sparse regions \cite{stanojevic2018robust, chen2016city}; (ii) \textit{Trajectory-merging-based} methods incrementally match new trajectories to existing roads while adding unmatched segments as new roads, yet they remain vulnerable to noisy inputs and frequently produce topologically inconsistent results \cite{he2018roadrunner, fathi2010detecting, karagiorgou2012vehicle}; (iii) \textit{Kernel Density Estimation-based} methods convert trajectories into continuous density fields for road extraction, but their rasterization process inevitably loses local geometric details and trajectory context, leading to systematic deviations from actual road topologies \cite{biagioni2012map, davies2006scalable, dey2017improved}.

\begin{figure*}[t]
  \centering
  \includegraphics[width=\linewidth]{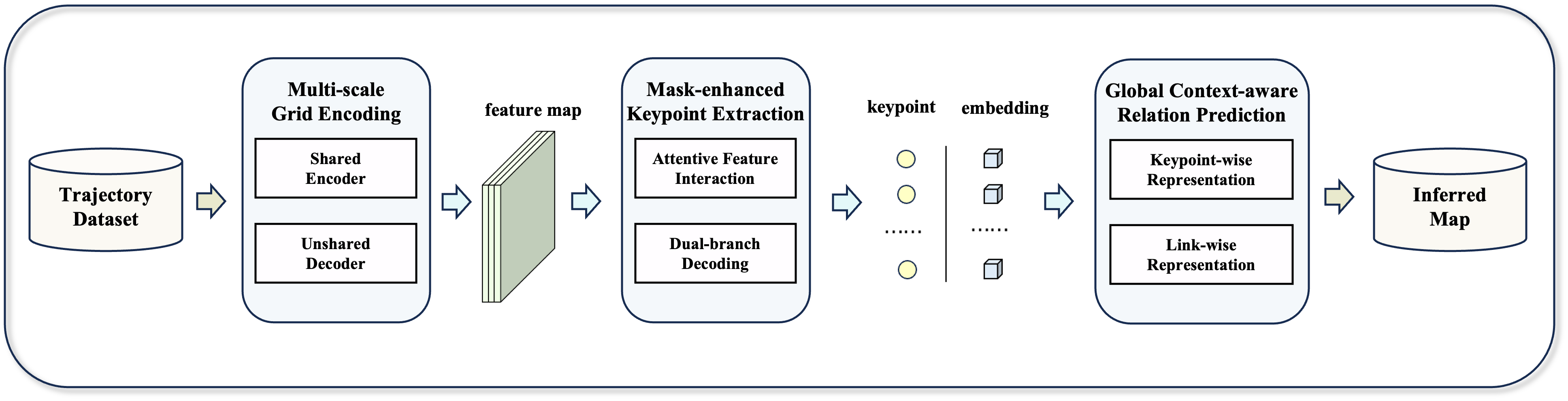}
  \caption{overall framework}
  \Description{description.}
\end{figure*}

\subsection{Learning-based Map Inference Methods}
Supervised learning-based map inference methods predominantly adopt three distinct methodological frameworks: (i)\textit{Segmentation-based} approaches formulate road extraction as a semantic segmentation problem using spatiotemporal features like trajectory density and directional coherence, yet their performance degrades significantly with trajectory noise and often yields incorrect topological connections in data-sparse regions. These methods also suffer from post-processing dependencies for morphological refinement, compromising both computational efficiency and topological accuracy \cite{eftelioglu2022ring, ruan2020learning}; (ii) \textit{Intersection linking-based} methods sequentially detect junctions and interpolate centerlines, but their performance critically depends on junction localization precision, as errors propagate through the linking phase and distort the final network topology \cite{shen2023sami}; (iii) \textit{Incremental-based} methods construct road networks iteratively through keypoint prediction, though they are prone to two fundamental issues: accumulated prediction errors causing path deviation, and initialization sensitivity resulting in unstable network generation \cite{bastani2018roadtracer}.

\section{Methods}

\subsection{Problem Statement}

\textbf{Definition 1} (Trajectory). A trajectory $tr = \{p_1, p_2, ... , p_n\} $ is a temporally ordered sequence of spatiotemporal points, where each point $p_i = (lng_i, lat_i, t_i)$ encodes a geospatial coordinate $(lng_i, lat_i)$ with an associated timestamp $t_i$, satisfying the temporal ordering constraint  $t_i \leq t_j$ for all $1\leq i \leq j \leq n $.

\textbf{Definition 2 (Road Map).} A road map is formally represented as a directed graph $G = <V, E>$, where:
\begin{itemize}
    \item $V$ is the set of \emph{key points} encoding road intersections and road shape points, each associated with a unique geospatial coordinate $v=(lng, lat)$;
    \item  $E \subseteq V \times V$ defines the set of directed edges, where an edge $e=<u, v>$ denotes a directly traversable road segment from \emph{key point} $\textit{u}$ to \emph{key point} $\textit{v}$.
\end{itemize}

\textbf{Problem Statement.} Given an observed trajectory dataset $T=\{tr_1, tr_2,…,tr_n\} $ within a bounded geographic region, the map inference problem requires the reconstruction of a directed graph $G = <V, E>$ representing latent road network that optimally explains $T$ under the constraints of completeness and physical plausibility.

\subsection{Solution Overview}

The proposed framework comprises three core components, as illustrated in Fig. 2. First, the \emph{multi-scale grid encoding} module discretizes the target region into grid cells and extracts hierarchical spatiotemporal features using a \emph{Deep Layer Aggregation} (\emph{DLA}) model. Second, the \emph{mask-enhanced keypoint extraction} module employs a \emph{dual-decoder} architecture that combines local geometric features with global semantic context from parallel segmentation, addressing road fragmentation in trajectory-sparse regions through complementary feature fusion. Third, the \emph{global context-aware relation prediction} module leverages both \emph{keypoint-level} and \emph{link-level} contextual information to accurately infer topological connections.


\subsection{Multi-scale Grid Encoding}
\begin{figure}[!htbp]
  \centering
  \includegraphics[width=\linewidth]{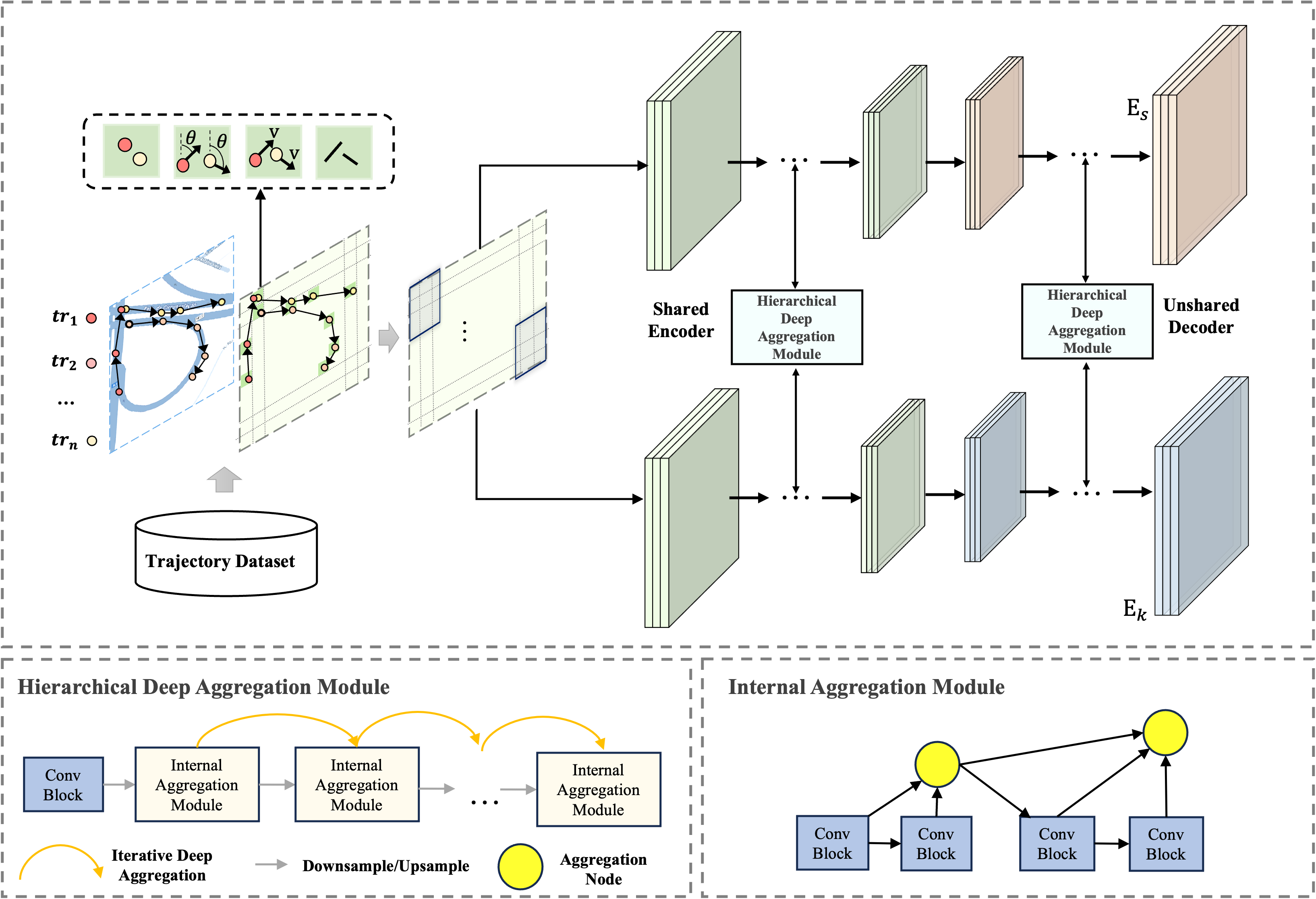}
  \caption{Multi-scale Grid Encoding}
  \Description{trajectory representation based on dla}
\end{figure}

Building upon \cite{ruan2020learning}, we formulate map inference as an image processing task by first rasterizing the road network into an $M \times N$ grid with $1-meter$ resolution. For each grid cell, we extract four behavioral features: (i) \emph{Point Frequency}, counting trajectory points within the cell; (ii) \emph{Point Direction Distribution}, computed by mapping the angular direction between consecutive points into eight discrete bins; (iii) \emph{Average Speed}, derived from the mean Euclidean distance between consecutive points normalized by time intervals; and (iv) \emph{Line Frequency}, enumerating trajectory segments connecting consecutive points. These features are organized into a grid feature matrix, which is subsequently partitioned into non-overlapping tiles following \cite{eftelioglu2022ring} to accommodate standard image processing constraints. Each tile is processed as an image with grid cells corresponding to pixels, enabling multi-scale feature extraction through the \emph{DLA} model\cite{yu2018deep}.

\textbf{Deep Layer Aggregation-based Multi-scale Grid Representation.} Road networks exhibit inherent scale variations, where major roads with greater width and trajectory density demonstrate distinct feature characteristics compared to narrower side streets with sparser trajectory coverage. This disparity complicates consistent road identification. To address this, we develop a grid encoding module based on the \emph{DLA} that effectively integrates multi-scale features. As illustrated in Fig. 3, this approach is inspired by proven multi-scale learning approaches in object detection. Unlike conventional feature extractors like \emph{FCN} and \emph{U-Net}, \emph{DLA} implementation employs hierarchical feature aggregation to optimize multi-scale feature fusion, significantly enhancing the model's capacity to recognize roads across different scales.

\subsection{Mask-enhanced Keypoint Extraction}
\begin{figure}[h]
  \centering
  \includegraphics[width=\linewidth]{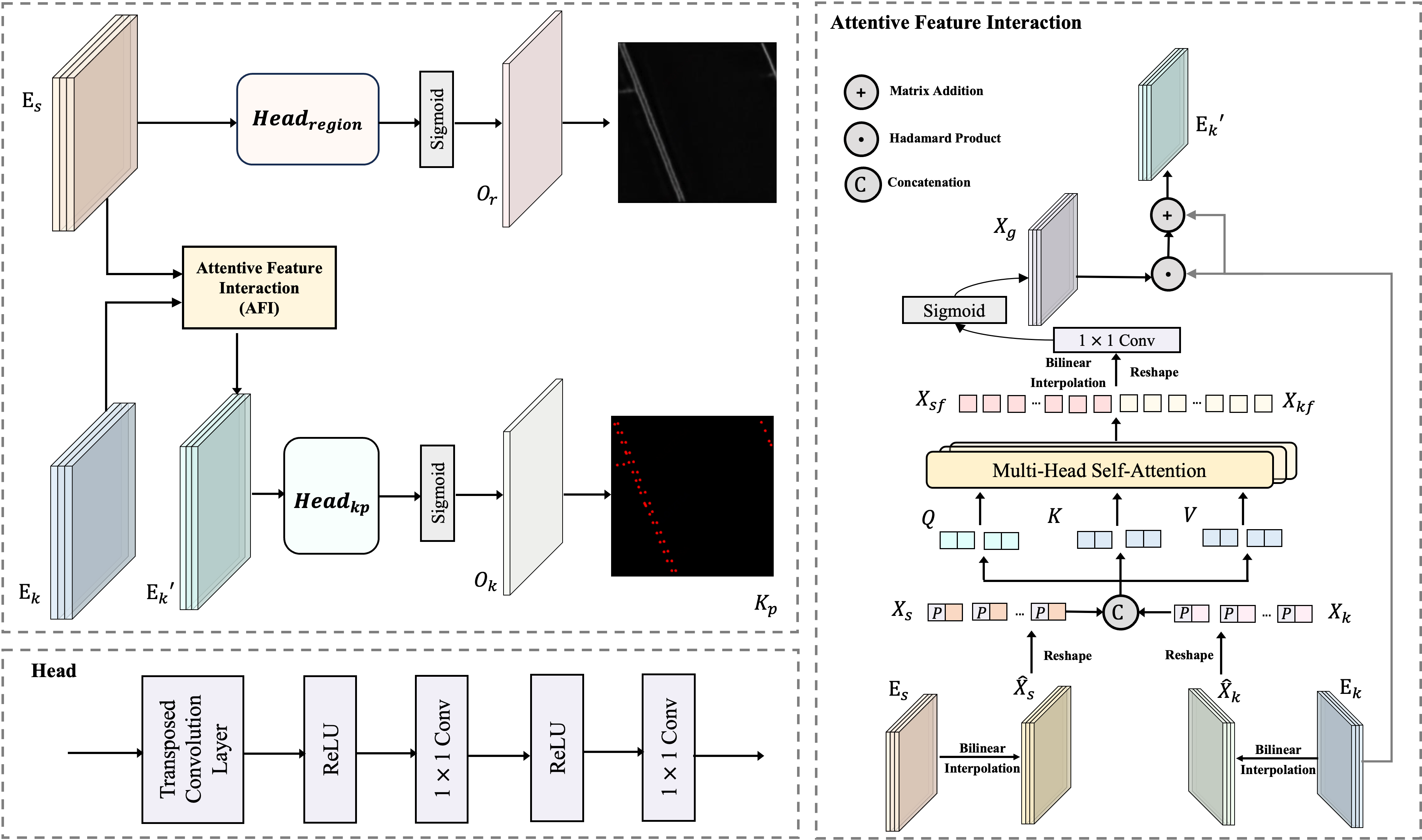}
  \caption{Mask-enhanced Keypoint Extraction}
  \Description{description.}
\end{figure}

Trajectory density images exhibit fundamentally different characteristics from satellite imagery, with spatially non-uniform distributions that often result in sparse point coverage along certain road segments. This irregularity causes existing keypoint extraction methods to produce fragmented road networks. As shown in Fig. 4, our solution introduces a \emph{Mask-enhanced Dual Decoder} (\emph{MeDD}) that jointly performs road region segmentation and keypoint extraction through an attentive feature interaction mechanism. The module's efficacy stems from complementary representations: while keypoint extraction identifies discrete critical locations, road segmentation provides comprehensive per-grid spatial context through dense prediction. Specifically, \emph{MeDD} employs an \emph{Attentive Feature Interaction Module} (\emph{AFIM}) to enrich keypoint-oriented representations ($\textbf{E}_{k}$) with segmentation-derived features ($\textbf{E}_{s}$), then simultaneously predicts both road keypoints and region masks through parallel decoding branches.

\textbf{Attentive Feature Interaction}. This step aims to enhance the keypoint-oriented grid representation $\textbf{E}_{k}$ by incorporating structural guidance from the segmentation-oriented grid representation $\textbf{E}_{s}$. First, both $\textbf{E}_{s}$ and $\textbf{E}_{k}$ are upsampled via bilinear interpolation to obtain patch matrices $\hat{\textbf{X}}_s \in \mathbb{R}^{l \times l \times C}$ and $\hat{\textbf{X}}_k \in \mathbb{R}^{l \times l \times C}$, where $l$ denotes the preset resolution. These matrices are then reshaped into patch embedding sequences $\textbf{X}_s \in \mathbb{R}^{2l^2 \times C}$ and $\textbf{X}_k \in \mathbb{R}^{2l^2 \times C}$, respectively. Positional embeddings are added to both sequences to preserve spatial information, resulting in $\textbf{X}_s = \textbf{X}_s + PE(\textbf{X}_s)$ and $\textbf{X}_k = \textbf{X}_k + PE(\textbf{X}_k)$, following the approach in \cite{vaswani2017attention}. The two sequences are concatenated and fed into a multi-head self-attention layer to generate fused representations $\textbf{X}_{sf} \in \mathbb{R}^{l^2 \times C}$ and $\textbf{X}_{kf} \in \mathbb{R}^{l^2 \times C}$. These are then concatenated along the channel dimension, reshaped, and upsampled through bilinear interpolation to produce the enhanced feature map $\textbf{E}_{g} \in \mathbb{R}^{H \times W \times C}$. A convolution block followed by a sigmoid activation is applied to generate the guidance map $\textbf{X}_{g} \in \mathbb{R}^{l \times l \times C}$. Finally, the enhanced keypoint-oriented grid representation $\textbf{E}_{ek}$ is obtained by performing a Hadamard product between $\textbf{X}_{g}$ and the original $\textbf{E}_{k}$, completing the enhancement process. This operation can be formulated as:
\begin{equation}
  \textbf{X}_{sf}, \textbf{X}_{kf}=MHSA(\textbf{X}_f)
\end{equation}
\begin{equation}
  \textbf{E}_{g} = (Us(Reshape(\textbf{X}_{sf})) \| Us(Reshape(\textbf{X}_{sf})))
\end{equation}
\begin{equation}
  \textbf{X}_{g} =  \sigma(Conv(\textbf{E}_{g}))
\end{equation}
\begin{equation}
  \textbf{E}_{ek} = \textbf{E}_{k} + \textbf{X}_{g} \odot \textbf{E}_{k} 
\end{equation}
where $MHSA$ denotes the multi-head self-attention, $Us$ denotes the upsampling operation, $Conv$ refers to a $1 \times 1$ convolution layer followed by a \emph{LeakyReLU} activation function, $\sigma$ denotes the sigmoid function and $\odot$ signifies the \emph{Hadamard product} (element-wise multiplication).

\textbf{Dual-branch Decoding}. Building on the enhanced keypoint-oriented grid representation $\textbf{E}_{k}^{'}$ and the segmentation-oriented grid representation $\textbf{E}_{s}$, the dual-branch decoder predicts the road keypoint probability map $O_k \in \mathbb{R}^{H \times W \times 1}$ and the road region mask $O_r \in \mathbb{R}^{H \times W \times 1}$ using two separate task heads. Each head shares the same architectural structure, consisting of a transposed convolution layer followed by a \emph{ReLU} activation, a $1\times1$ convolutional layer with another \emph{ReLU} activation, and finally a $1 \times 1$ convolutional layer to produce the output. It is important to note that the two heads are not parameter-shared. The overall process can be formulated as:
\begin{equation}
    O_k = \sigma(Head_{kp}(\textbf{E}_{k}^{'}))
\end{equation}
\begin{equation}
    O_r = \sigma(Head_{region}(\textbf{E}_{s}))
\end{equation}
where $Head_{kp}$ denotes the keypoint prediction head, and $Head_{region}$ denotes the road region segmentation head, with $\sigma$ representing the sigmoid function. Following \cite{zhang2019ppgnet}, we apply the gaussian filter to smooth the keypoint probability map, and the points with locally maximal probabilities are extracted as the final detected keypoints $Kp$.

\textbf{Model Learning}. The keypoint extraction module is trained using a combined objective function that integrates the keypoint estimation loss and the road region segmentation loss, which can be formulated as:
\begin{equation}
    L_{kp} = \sum_{(x,y) \in [1..W]\times[1..H]}L_{ce}(O_{k}^{x,y}, \hat{O}_{k}^{x,y}) + \lambda_1 L_{ce}(O_r^{x,y}, \hat{O}_{r}^{x,y})
\end{equation}
where the former term represents the keypoint estimation loss, and the latter corresponds to the road region segmentation loss. $L_{ce}$ denotes the cross entropy loss function, $\hat{O_k}$ and $\hat{O_r}$ represent the ground truth keypoint probability map and road region mask, respectively, and $\lambda_1$ is a hyperparameter that balances the contribution of the two loss components.

\subsection{Global Context-aware Relation Prediction}
\begin{figure}[h]
  \centering
  \includegraphics[width=\linewidth]{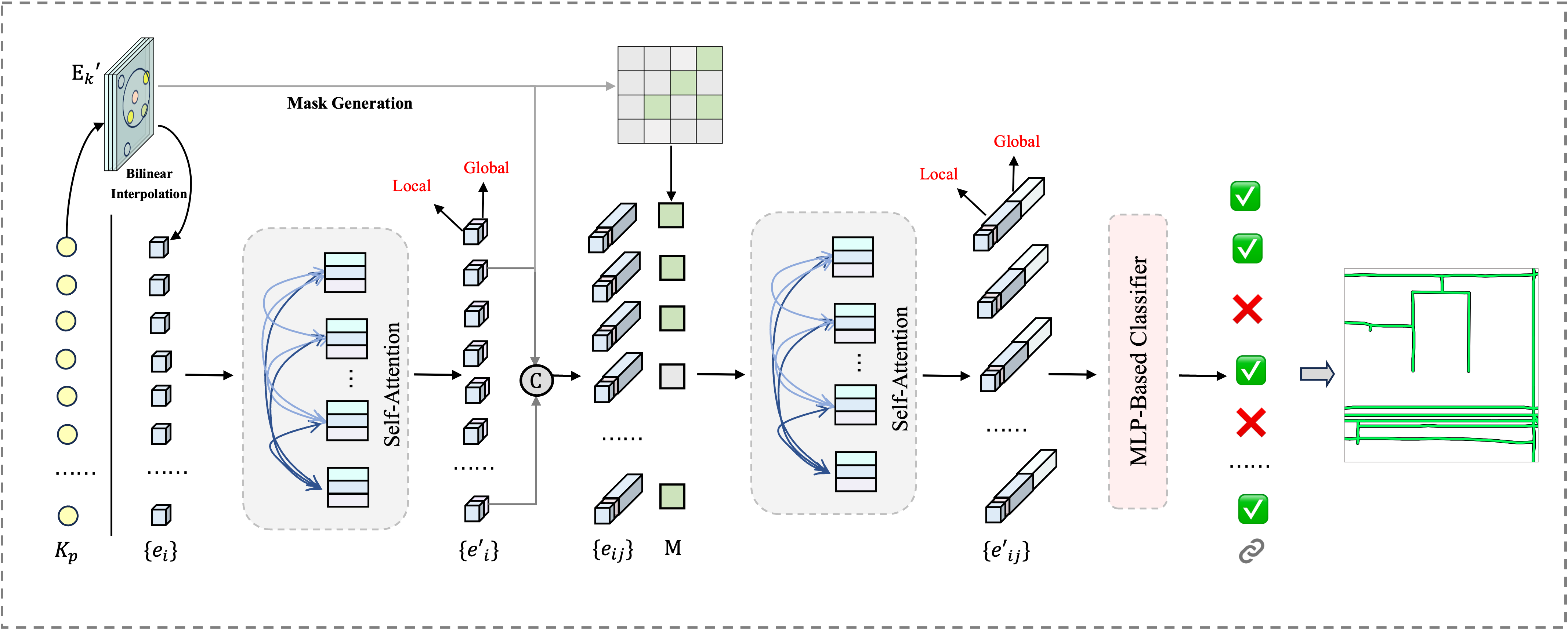}
  \caption{Global Context-aware Relation prediction}
  \Description{description.}
\end{figure}

The relation prediction task determines connectivity between keypoints to reconstruct road networks. Current approaches predominantly rely on either heuristic rules using distance thresholds and trajectory frequency, or \emph{MLP}-based local prediction networks\cite{he2022td, yang2023topdig, ding2023pivotnet}. These methods suffer from a critical limitation: their exclusive focus on local keypoint neighborhoods fails to capture essential global road characteristics, particularly the inherent linear structures and contextual dependencies among connected keypoints. As illustrated in Fig. 5, we develop a \emph{Global Context-aware Relation Prediction} module that integrates both keypoint-level and link-level global contexts, augmented by a redundancy-aware optimization strategy to enhance prediction accuracy.

\textbf{Keypoint-wise Representation Enrichment}. Given the fetched keypoint set $Kp$, we first employ bilinear interpolation to extract the keypoint representations from the enhanced keypoint-oriented grid representations $E_{ek}$. Then, we input all keypoint representations $\{e_{i} \in \mathbb{R}^{C}| i \in Kp\}$ with their positional embeddings to a self-attention layer to capture the global context between the keypoints. The output keypoint representations from the self-attention are concatenated with $\{e_{i}\}$ to form the final keypoint representations $\{e_{i}^{'}\}$. 

\textbf{Link-wise Representation Enrichment. Each ordered pair of any two keypoints is considered as a candidate link pair. Thus, we first generate representations of all candidate link pairs $\{e_{ij}=e_{i}^{'}||e_{j}^{'}||e_{ij}^{link} | \forall i, j \in Kp, i \ne j\}$ by concatenating the two keypoint representations and their corresponding link embedding. Here the link embedding $e_{ij}^{link}$ is obtained by performing average pooling over the grid representations along the line segment connecting keypoint $i$ and $j$. Then, we input them into an another self-attention layer to capture the global context between the candidate link pairs. Note that the self-attention mechanism entails high computational overhead when applied to a large set of keypoints, due to its time complexity being $\mathcal{O}(n^2)$, where $n$ denotes the number of keypoints. To alleviate this, we introduce a masking mechanism. Given a candidate link pair $(i, j)$, we consider another link $(k, l)$ in the attention calculation only if keypoint $k$ or $l$ lies with in a $d_l$-meter radius of either $i$ or $j$. Thus, the mask matrix is formulated as:
\begin{equation}
M_{(i,j), (k,l)} = 
\begin{cases}
0 & \text{if } \exists u \in \{i, j\}, v \in \{k, l\}, \text{ s.t. } dis(u, v) \leq d_l \\
1 & \text{otherwise}
\end{cases}
\end{equation}
where $dis(\cdot)$ denotes the euclidean distance function. Through incorporating the masking mechanism, the complexity decreases from $\mathcal{O}(n^2)$ to $\mathcal{O}(mn)$, where $m \ll n$ denotes the number of keypoints that satisfied the constraint. The final link pair representations $\{e_{ij}^{'}\}$ are fetched by concatenating the original link pair representations with the output of the self-attention layer.
}

Finally, we employ an MLP-based classifier to judge whether a candidate link pair is valid to obtain the road map. 

\textbf{Redundancy-aware Model Learning}. The most straightforward strategy to train GCRP is by extracting all valid link pairs from the ground truth map as positive samples and viewing the other link pairs as the negative ones. However, the excessive number of negative samples poses a significant class imbalance challenge. Considering that two keypoints that are closer to each other are more likely to be identified as a redundant link, we design a hard negative sampling strategy to effectively train the GCRP. Specifically, for a ground truth keypoint $\hat{kp}_i$, we consider it to form a positive sample $(\hat{kp}_i, \hat{kp}_j)$ with another keypoint $\hat{kp}_j$ if $\hat{kp}_j$ is a 1 hop neighbor of $\hat{kp}_i$, and to form a hard negative sample $(\hat{kp}_i, \hat{kp}_j)$ if $\hat{kp}_j$ is a 2-to-$N_r$ hop neighbor of $\hat{kp}_i$, or the euclidean distance between $\hat{kp}_i$ and $\hat{kp}_j$ is less than $3d_{kp}$. Through this sampling strategy, we have extracted the positive samples and hard negative samples for each ground keypoint $\hat{kp}_i$. Then we train the GCRP module using the binary cross-entropy loss, denoted as $L_{rel}$, with reduction performed by averaging over the number of keypoints.

Therefore, the overall loss function to train DGMap is formulated as:
\begin{equation}
    L = L_{kp} + \lambda_2 L_{rel}
\end{equation}
where $\lambda_2$ denotes the hyper-parameter to control the weights of two losses.

\section{Experiments}
\subsection{Experimental Settings}

\begin{figure*}[h]
  \centering
  \includegraphics[width=\linewidth]{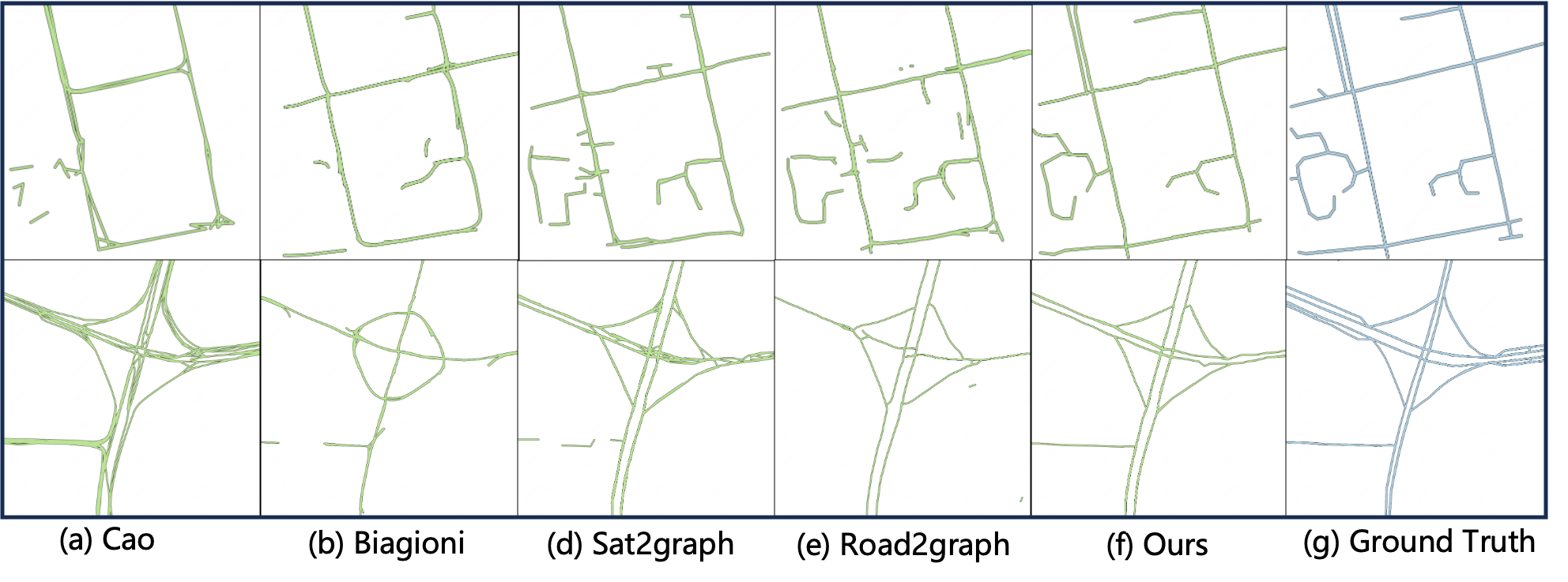}
  \caption{Visualization of inferred road maps}
  \Description{description.}
\end{figure*}

\textbf{Datasets.} This study utilizes three trajectory datasets (\textbf{BJ24}, \textbf{SZ24}, and \textbf{WX20}) for experimental validation and analysis. 
\textbf{BJ24} consists of taxi trajectory data collected from the \textit{DidiChuxing} platform in Beijing over a seven-day period.
\textbf{SZ24} contains similar taxi trajectory data gathered from Shenzhen area during the same duration.
\textbf{WX20}$\footnote{\url{https://pan.baidu.com/s/1uz8nU5Ztp87QNnOTfV4iiQ?pwd=onhs}}$ is a publicly available taxi trajectory dataset covering the city of Wuxi.
All datasets include four fundamental attributes: Trajectory identifier, Timestamp, Latitude, and Longitude. The ground truth road networks for these regions were obtained from \emph{OpenStreetMap}, a crowd-sourced mapping platform that provides collaboratively maintained, globally available road network data.

\begin{table}[!htbp]
\centering
\caption{DESCRIPTION OF EXPERIMENTAL DATASETS}\label{tab:aStrangeTable}
\begin{tabular}{cccc}
\toprule
Dataset & BJ24 & SZ24 & WX20 \\ \hline
Time Span & 240108--240115 & 240520--240526 & 200718--200801 \\
Sampling Rate (s) & 5 & 5 & 3 \\
Size (km$^2$) & $15 \times 17$ ($3 \times 3$) & $19 \times 21$ ($4 \times 4$) & $27 \times 31$ ($6 \times 6$) \\
Point Count & 1.7M & 2.3M & 3.6M \\
\bottomrule
\end{tabular}
\end{table}

\textbf{Baselines.} We compare \emph{DGMap} with five  trajectory-based map inference baselines: 
\emph{Cao} \cite{cao2009gps} proposes an automated road network generation framework that combines trajectory smoothing with optimized clustering.
\emph{Biagioni} \cite{biagioni2012map} generates road networks by first applying kernel density estimation to trajectory points, followed by a map matching algorithm to eliminate low-confidence edges. 
\emph{Kharita} \cite{stanojevic2018robust} formulates map inference as a network alignment problem: it infers vertices via \emph{K-Means} clustering, connects them based on passing trajectories.
\emph{RingNet} \cite{eftelioglu2022ring}  an improved version of \emph{DeepMG} \cite{ruan2020learning}, enhances centerline mask prediction through self-attention in \emph{D-LinkNet}.
\emph{Sami} \cite{shen2023sami} detects intersections by analyzing behavioral differences and connects them using a \emph{shape-aware} centerline fitting method for road curve generation.

By treating each input tile as a computer vision image through the \emph{Grid Feature Extraction} module, we also evaluate \emph{DGMap} against methods designed for road network inference from satellite imagery to assess its robustness:
\emph{Sat2Graph} \cite{he2020sat2graph} introduces a graph encoding approach for road networks, where a \emph{DLA} network infers a \emph{17-dimensional} graph code tensor, which is then decoded into a topological structure via a handcrafted decoding module.
\emph{TDRoad} \cite{he2022td} detects road keypoints using a  \emph{DLA} network and employs an \emph{MLP}-based connectivity model to predict pairwise links, forming the final road network.
\emph{Road2Graph} \cite{zao2024road} proposes a graph encoding method that integrates multi-scale contextual information by leveraging transformer-based encoding of deep convolutional features.

\textbf{Evaluation metrics}. In the evaluation, we adopt \textit{TOPO} and \textit{Average Path Length Similarity} to assess the accuracy of the inferred road networks, as these metrics are robust to structural errors and provide reliable performance evaluation \cite{chao2020survey}.
\emph{TOPO} \cite{biagioni2012map} evaluates the similarity between the inferred and real road networks by comparing the \emph{precision}, \emph{recall}, and \emph{F1-score} of reachable subgraphs for each vertex.
\emph{Average Path Length Similarity (APLS)} \cite{mattyus2017deeproadmapper} measures the difference in shortest path lengths between source-target point pairs mapped onto both the inferred and ground truth road networks.

\textbf{Implementation details.} We implement \emph{DGMap} using PyTorch 1.12.0 and conduct all experiments on an NVIDIA RTX A6000. The height and width of each tile are fixed at 256 pixels. For \emph{BJ24}, \emph{SZ24}, and \emph{WX20}, the numbers of training, validation, and test samples are (1813, 518, 259), (1671, 478, 239) and (1723, 492, 24) respectively. During training, we use the Adam optimizer with an initial learning rate of 0.001, which is reduced by a factor of 10 every 10 epochs.

\begin{table*}[t]
\caption{PERFORMANCE EVALUATION FOR DIFFERENT METHODS ON MAP INFERENCE TASK}
\setlength{\tabcolsep}{2 pt}
\begin{tabular}{c|c|cccc|cccc|cccc} 
\hline
\textbf{Type}& \textbf{Methods}& \multicolumn{4}{c|}{\textbf{BJ24}}
& \multicolumn{4}{c|}{\textbf{SZ24}}  
& \multicolumn{4}{c}{\textbf{WX20}} \\ \cline{3-14}
& & \multicolumn{1}{l}{\textbf{Precision}} & \multicolumn{1}{l}{\textbf{Recall}} 
& \multicolumn{1}{l|}{\textbf{F1-score}} & \multicolumn{1}{l|}{\textbf{APLS}}
& \multicolumn{1}{l}{\textbf{Precision}} & \multicolumn{1}{l}{\textbf{Recall}} 
& \multicolumn{1}{l|}{\textbf{F1-score}} & \multicolumn{1}{l|}{\textbf{APLS}}
& \multicolumn{1}{l}{\textbf{Precision}} & \multicolumn{1}{l}{\textbf{Recall}} 
& \multicolumn{1}{l|}{\textbf{F1-score}} & {\textbf{APLS}}     \\ \hline

\multirow{4}{*}{Non-Learning}
& Cao09
& 0.7183& 0.1837&  \multicolumn{1}{c|}{0.2926} & \multicolumn{1}{c|}{0.2591} 
& 0.6953& 0.1907&  \multicolumn{1}{c|}{0.2993} & \multicolumn{1}{c|}{0.2713} 
& 0.6937& 0.1943&  \multicolumn{1}{c|}{0.3036} & 0.3258\\
& Biagioni12
& 0.7924& 0.2093&  \multicolumn{1}{c|}{0.3311} & \multicolumn{1}{c|}{0.2316} 
& 0.7329& 0.2448&  \multicolumn{1}{c|}{0.3670} & \multicolumn{1}{c|}{0.2591} 
& 0.7549& 0.2546&  \multicolumn{1}{c|}{0.3808} & 0.2037\\
& Kharita18
& 0.7328& 0.1907&  \multicolumn{1}{c|}{0.3026} & \multicolumn{1}{c|}{0.2773} 
& 0.6995& 0.2203&  \multicolumn{1}{c|}{0.3350} & \multicolumn{1}{c|}{0.2829} 
& 0.7146& 0.2193&  \multicolumn{1}{c|}{0.3356} & 0.2628\\
& Sami23& 0.8283& 0.3051&  \multicolumn{1}{c|}{0.4459} & \multicolumn{1}{c|}{0.2979} 
& 0.8375& 0.2987&  \multicolumn{1}{c|}{0.4403} & \multicolumn{1}{c|}{0.3262} 
& 0.8245& 0.3043&  \multicolumn{1}{c|}{0.4445} & 0.3546\\ \hline
\multirow{2}{*}{Learning}
& Ringnet22& 0.8493& 0.2892&  \multicolumn{1}{c|}{0.4315} & \multicolumn{1}{c|}{0.3318} 
& 0.8648& 0.3007&  \multicolumn{1}{c|}{0.4462} & \multicolumn{1}{c|}{0.4023} 
& 0.8576& 0.3029&  \multicolumn{1}{c|}{0.4448} & 0.4256\\ 
& DGMap& \textbf{0.8699}& \textbf{0.3212}&  \multicolumn{1}{c|}{\textbf{0.4691}} & \multicolumn{1}{c|}{\textbf{0.4648}} 
& \textbf{0.8862}& \textbf{0.3163}&  \multicolumn{1}{c|}{\textbf{0.4662}} & \multicolumn{1}{c|}{\textbf{0.4982}} 
& \textbf{0.8764}& \textbf{0.3328}&  \multicolumn{1}{c|}{\textbf{0.4824}} & \textbf{0.5583}\\ \hline
\end{tabular}
\label{main results}
\end{table*}

\subsection{Performance Comparison} 

\textbf{Quantitative Comparison.} Table 2 presents the quantitative results of various map inference methods, with bold values indicating top performance. Key observations can be summarized as follows:
\begin{itemize}
    \item Our proposal achieves state-of-the-art performance on both \emph{APLS} and \emph{TOPO-F1} metrics across three real-world datasets: \emph{BJ24}, \emph{SZ24}, and \emph{WX24}. Specifically, it surpasses the best baseline by 3.86\%, 2.00\%, and 1.87\% in \emph{F1-score}, and by 9.97\%, 8.44\%, and 12.26\% in \emph{APLS}, respectively. 
    \item Learning-based methods generally demonstrate superior overall performance compared to non-learning-based approaches, primarily due to their ability to utilize existing road networks as supervisory signals during model training. 
\end{itemize}
\begin{table}[!htbp]
\caption{PEFORMANCE EVALUATION WITH METHODS IN THE FIELD OF MAP INFERENCE FROM SATELLITE IMAGES}
\setlength{\tabcolsep}{3 pt}
\begin{tabular}{c|c|cccc} 
\hline
\textbf{Dataset}
&\textbf{Methods}
&\textbf{Precision} 
&\textbf{Recall} 
& \multicolumn{1}{l|}{\textbf{F1-score}} 
&  \textbf{APLS}  \\ \hline
\multirow{4}{*}{\textbf{BJ24}}
& Sat2graph20  
& 0.8231& 0.3117& \multicolumn{1}{c|}{0.4521} & 0.3983\\
& TDRoad22     
& 0.8194& 0.2746& \multicolumn{1}{c|}{0.4113} & 0.3752\\
& Road2graph24 
& \textbf{0.8769}& 0.3013& \multicolumn{1}{c|}{0.4485} & 0.4109\\ \cline{2-6} 
& DGMap& 0.8699& \textbf{0.3212}& \multicolumn{1}{c|}{\textbf{0.4691}} & \textbf{0.4648}\\ \hline
\multirow{4}{*}{\textbf{SZ24}}
& Sat2graph20
& 0.8148& 0.3081& \multicolumn{1}{c|}{0.4471}  & 0.4173\\
& TDRoad22
& 0.8392& 0.2835& \multicolumn{1}{c|}{0.4238}  & 0.4237\\
& Road2graph24 
& 0.8813& 0.3068& \multicolumn{1}{c|}{0.4551}   & 0.4581\\ \cline{2-6} 
& DGMap& \textbf{0.8862}& \textbf{0.3163}& \multicolumn{1}{c|}{\textbf{0.4662}} & \textbf{0.4982}\\ \hline
\multirow{4}{*}{\textbf{WX20}}
& Sat2graph20  
& 0.8394& \textbf{0.3351}& \multicolumn{1}{c|}{0.4790} & 0.5214\\
& TDRoad22 
& 0.8145& 0.2735& \multicolumn{1}{c|}{0.4095} & 0.4948\\
& Road2graph24 
& \textbf{0.8779}& 0.2927& \multicolumn{1}{c|}{0.4390} & 0.5238\\ \cline{2-6} 
& DGMap& 0.8764& 0.3328& \multicolumn{1}{c|}{\textbf{0.4824}} & \textbf{0.5583}\\ \hline
\end{tabular}
\label{satellite}
\end{table}

\textbf{Comparison with Methods in the Fields of Map Inference from Satellite Images.} Through multi-scale grid encoding, each geospatial tile is converted into an image-like representation, allowing the application of road network extraction techniques originally designed for satellite imagery. Comparative experiments with \emph{Sat2Graph}, \emph{TDRoad}, and \emph{Road2Graph} validate the effectiveness and generalization ability of our \emph{DGMap}. As presented in Table 3, \emph{Road2Graph} achieves better performance than \emph{Sat2Graph} and \emph{TDRoad} across all datasets, highlighting the benefits of incorporating global multi-scale topological context for road network modeling. The consistently higher \emph{APLS} and \emph{Topo-F1} scores obtained by \emph{DGMap} across all evaluation datasets further demonstrate its robustness and superior mapping capability.

\textbf{Visualization Comparison.} We further present visual comparisons on two representative regions across three real-world trajectory datasets. \emph{Region 1} represents areas with sparse trajectory coverage, while \emph{Region 2} corresponds to high-density scenarios involving complex intersections and parallel roads. As shown in Figure 6, the inferred road networks are depicted as green solid lines, and the ground-truth road networks are represented by blue solid lines. In \emph{Region 1}, existing methods struggle to reconstruct complete and topologically accurate road structures due to insufficient trajectory data, often producing fragmented outputs that degrade network connectivity. In contrast, in \emph{Region 2} with dense trajectory distribution, these approaches tend to generate excessive redundant segments, especially around intersections and parallel roads. Our method outperforms existing approaches in both scenarios, delivering more accurate, coherent, and robust road network reconstructions.

\subsection{Model Analysis} 
\textbf{Ablation study.} To evaluate the contribution of each component in \emph{DGMap}, we conduct ablation studies using five variants on the \emph{BJ24} dataset: i) \textbf{w/o DR} replaces the \textit{Deep Layer Aggregation-based Multi-scale Grid Representation} with a U-Net; ii) \textbf{w/o DD} removes the Dual-branch Decoding and uses a single decoder with shared parameters; iii) \textbf{w/o AI} eliminates the \emph{Attentive Feature Interaction} module and instead directly concatenates the keypoint-oriented and segmentation-oriented grid representations; iv) \textbf{w/o KE} removes the \emph{Keypoint-wise Representation Enrichment}; v) \textbf{w/o LE} removes the \textit{Link-wise Representation Enrichment}. These ablation experiments provide insights into the role and effectiveness of each design choice in the overall architecture.
\begin{table}[!htbp]
\caption{ABLATION STUDY ON THREE DATASETS}
\setlength{\tabcolsep}{3 pt}
\begin{tabular}{c|cccc} 
\hline
\textbf{Methods} &\textbf{Precision} &\textbf{Recall} 
& \multicolumn{1}{l|}{\textbf{F1-score}} & \textbf{APLS}  \\ \hline
 w/o DR& 0.8412& 0.3048& \multicolumn{1}{c|}{0.4475} & 0.4507\\
 w/o AI& 0.8324& 0.2982& \multicolumn{1}{c|}{0.4391} & 0.4431\\
 w/o DD& 0.8319& 0.2871& \multicolumn{1}{c|}{0.4269} & 0.4472\\
  w/o KE& 0.8211& 0.3048& \multicolumn{1}{c|}{0.4446} & 0.4251\\
  w/o LE& 0.8296& 0.3157& \multicolumn{1}{c|}{0.4574} & 0.4328\\ \hline 
  DGMap& \textbf{0.8699}& \textbf{0.3212}& \multicolumn{1}{c|}{\textbf{0.4691}} & \textbf{0.4648}\\ \hline
\end{tabular}
\label{ablation}
\end{table}

\begin{itemize}
    \item \textbf{w/o DR} variant achieves lower performance than \emph{DGMap} in both \emph{F1-score} and \emph{APLS}, indicating that the \emph{Deep Layer Aggregation} (\emph{DLA}) module improves multi-scale feature fusion through hierarchical aggregation, thereby enhancing model robustness in environments with imbalanced trajectory density distributions.
    \item The \textbf{w/o AI} and \textbf{w/o DD} variants exhibit reduced recall and precision compared to \emph{DGMap}, demonstrating that the integration of global semantic and local topological features by the \emph{Attentive Feature Interaction} and \emph{Dual-branch Decoding} modules is critical for accurate road network inference, particularly in trajectory-sparse regions.
    \item Both \textbf{w/o KE} and \textbf{w/o LE} underperform \emph{DGMap} in terms of \emph{precision} and \emph{APLS}, confirming that modeling long-range dependencies between keypoints and capturing global connectivity patterns are essential for improving the model's ability to distinguish valid from spurious connections in the inferred road network.
\end{itemize}

\section{Conclusion}
This paper proposes an end-to-end keypoint-based framework for road map inference from trajectory data, addressing road fractures and road redundancy caused by uneven trajectory density distribution.
The method effectively mitigates these issues by integrating \emph{global semantic information extraction} with \emph{context-aware relational modeling}. Experiments on three real-world datasets demonstrate our method's significant improvements in accuracy and robustness over state-of-the-art approaches. Future work will explore the integration of multi-modal data, such as satellite imagery and remote sensing, to further enhance performance and overcome the limitations of trajectory-only inference.

\section{GenAI Usage Disclosure}
This work utilized AI-based tools exclusively for language editing and grammatical review.

\bibliographystyle{ACM-Reference-Format}
\balance
\bibliography{ref.bib}

\end{document}